%% file: paper.tex
\documentclass{article}
\usepackage[dvipsnames]{xcolor}
\usepackage[accepted]{icml2024}

\input{math_commands.tex}

\usepackage[utf8]{inputenc}
\usepackage[T1]{fontenc}
\usepackage{microtype}
\usepackage{graphicx}
\usepackage{booktabs,multirow}
\usepackage{algorithm,algorithmic}
\usepackage[font=small]{caption}
\usepackage{subcaption}
\usepackage{enumitem}
\setlist[itemize]{leftmargin=*}
\setlist[enumerate]{leftmargin=*}
\usepackage{wrapfig}

\usepackage{etoolbox} 
\makeatletter
\patchcmd{\BR@backref}{\newblock}{\newblock[page~}{}{}
\patchcmd{\BR@backref}{\par}{]\par}{}{}
\makeatother

\usepackage{amsmath}
\usepackage{amssymb}
\usepackage{mathtools}
\usepackage{amsthm}
\usepackage{xspace}

\colorlet{my-red}{BrickRed!90!Sepia}
\colorlet{my-blue}{Aquamarine!30!Blue}
\usepackage[backref=page]{hyperref}
\hypersetup{
  colorlinks,
  urlcolor  = my-red,
  linkcolor = my-blue,
  citecolor = my-blue,
}
\usepackage[capitalize,nameinlink,noabbrev]{cleveref}
\crefname{equation}{}{}
\Crefname{equation}{}{}

\usepackage{titlesec}
\titlespacing\section{0pt}{0pt plus 2pt minus 2pt}{0pt plus 2pt minus 2pt}

\theoremstyle{plain}

\theoremstyle{definition}

\theoremstyle{remark}

\icmltitlerunning{AutoFT: Learning an Objective for Robust Fine-Tuning}

\begin{document}

\twocolumn[
\icmltitle{
AutoFT: Learning an Objective for Robust Fine-Tuning
}



\icmlsetsymbol{equal}{*}

\begin{icmlauthorlist}
\icmlauthor{Caroline Choi}{equal,stanford}
\icmlauthor{Yoonho Lee}{equal,stanford}
\icmlauthor{Annie Chen}{stanford}
\icmlauthor{Allan Zhou}{stanford}
\icmlauthor{Aditi Raghunathan}{cmu}
\icmlauthor{Chelsea Finn}{stanford}
\end{icmlauthorlist}

\icmlaffiliation{stanford}{Department of Computer Science, Stanford University, USA}
\icmlaffiliation{cmu}{Department of Computer Science, Carnegie Mellon University, USA}

\icmlcorrespondingauthor{Caroline Choi}{cchoi1@stanford.edu}
\icmlcorrespondingauthor{Yoonho Lee}{yoonho@stanford.edu}

\icmlkeywords{Machine Learning, ICML}

\vskip 0.3in
]



\printAffiliationsAndNotice{}  

\newcommand{\ours}{\textsc{AutoFT}}
\newcommand{\oursbf}{\textbf{\textsc{AutoFT}}}
\newcommand{\LA}{\text{LearnAlg}}
\newcommand{\Dtr}{D_{\text{tr}}}
\newcommand{\Dval}{D_{\text{val}}}
\newcommand{\Ptr}{\mathcal{P}_{\text{tr}}}
\newcommand{\Pval}{\mathcal{P}_{\text{val}}}
\newcommand{\Pood}{\mathcal{P}_{\text{ood}}}
\DeclarePairedDelimiterX{\infdivx}[2]{(}{)}{%
  #1\;\delimsize|\delimsize|\;#2%
} 
\newcommand{\kld}[2]{\ensuremath{D_{KL}\infdivx{#1}{#2}}\xspace}
\newcommand{\lowvar}{{\color{BrickRed} 10^{-3}}}
\newcommand{\highvar}{{\color{BlueViolet} 1.0}}

\begin{abstract}
\input{sections/abstract.tex}
\end{abstract}

\section{Introduction}


Foundation models have emerged as a powerful tool in machine learning, demonstrating unprecedented performance across a wide variety of data distributions ~\citep{radford2021learning, ilharco_gabriel_2021_5143773, jia2021scaling}.
By pre-training on large and diverse datasets, these models encode rich, common-sense knowledge that can be leveraged in downstream tasks to improve performance.
However, while fine-tuning on additional task-specific data improves performance on the target distribution, it degrades performance on different distributions.

\begin{figure}
\vspace{-1mm}
\centering
  \includegraphics[width=0.9\linewidth]{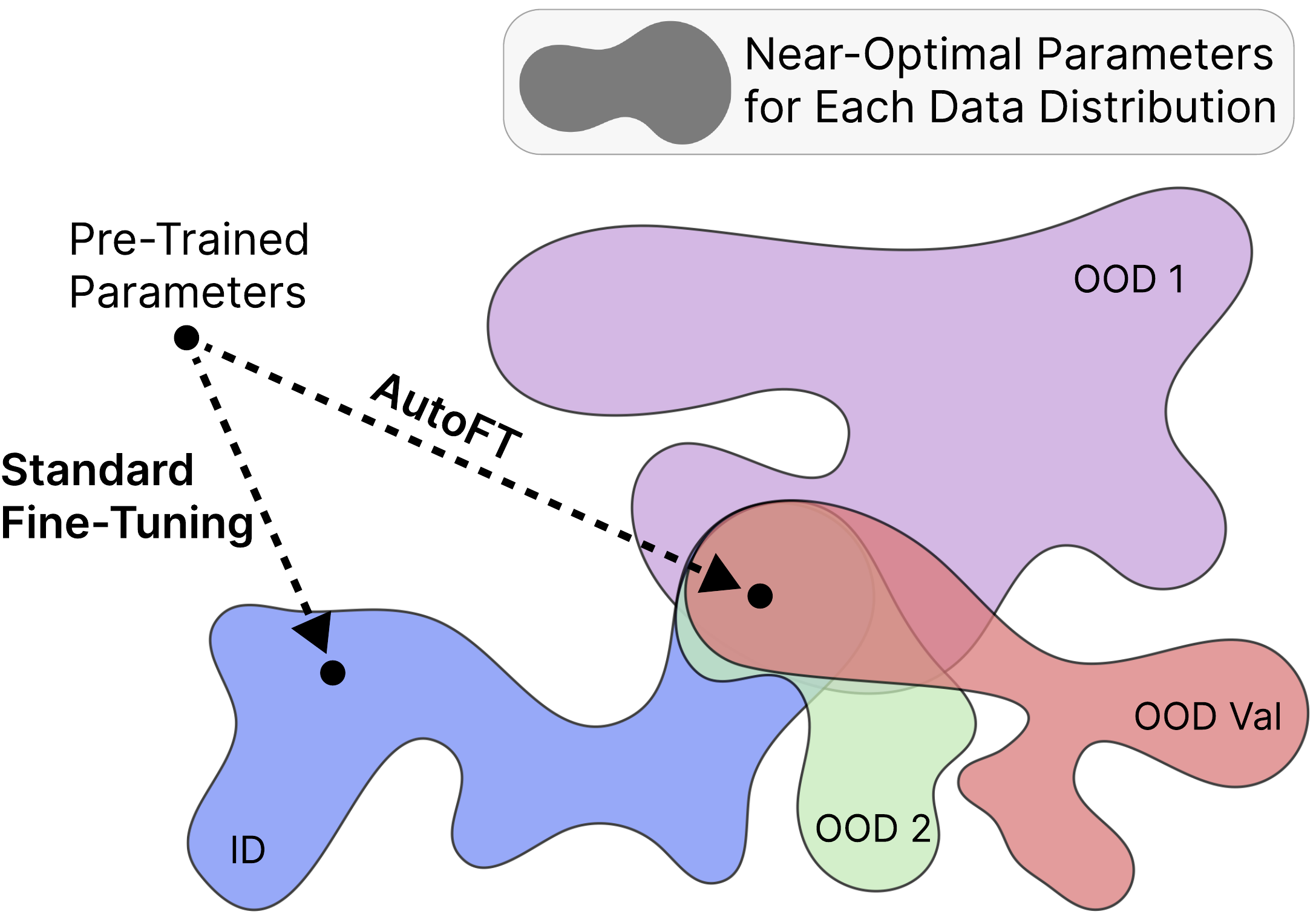}
  \caption{
  Overview of \ours{}: a method for robustly fine-tuning foundation models.
  While fine-tuning with in-distribution (ID) data (blue), \ours{} searches for a fine-tuning objective that maximizes performance on a small out-of-distribution validation set (red).
  This validation set serves as a proxy for performance on different distributions (green and purple), allowing \ours{} to learn a robust fine-tuning procedure.
  }
  \label{fig:overview}
\vspace{-2mm}
\end{figure}

\begin{figure*}
  \centering 
  \vspace{2mm}
  \includegraphics[width=.995\linewidth]{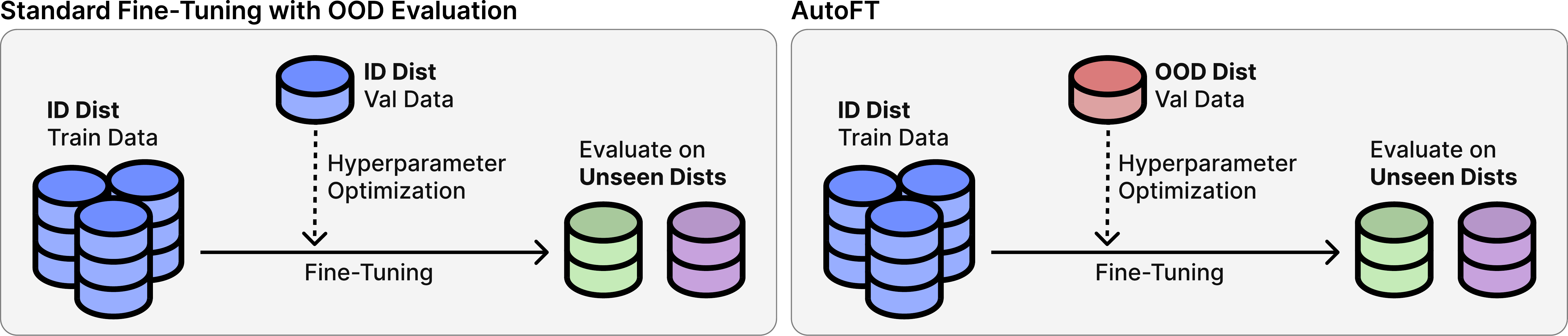}
  \vspace{-1mm}
  \caption{
  \label{fig:data_assumptions}
  A summary of our data assumptions and evaluation protocol.
  The standard approach is to optimize hyperparameters on a validation dataset drawn from the same distribution as the training data.
  In contrast, AutoFT employs a small out-of-distribution (OOD) validation set for hyperparameter optimization, enhancing the generalizability of the final model.
  We evaluate all fine-tuned models on data from unseen distribution shifts (green and purple).
  }
  \vspace{-1mm}
\end{figure*}

This issue has driven recent research on \textit{robust} fine-tuning, which aims to produce an adapted model that achieves good performance under distribution shifts. 
Prior works have proposed various regularization techniques to preserve the prior knowledge embedded in the foundation model, such as ensembling models before and after adaptation~\citep{wortsman2022robust} or initially fitting only the last layer~\citep{kumar2022fine}.
However, these approaches propose hard-coded fine-tuning procedures, which may not fully capture the intricate relationship between foundation model priors and task-specific data during adaptation.
Furthermore, this relationship can vary across fine-tuning tasks, and a one-size-fits-all approach may not account for these differences.

We introduce \ours{}, a novel approach for robust fine-tuning that aims to optimally balance between preserving foundation model priors and incorporating task-specific knowledge during adaptation.
Our key insight is that we can \textit{learn} the best adaptation strategy for a given fine-tuning task.
\ours{} learns a fine-tuning objective and hyperparameters using a small out-of-distribution (OOD) validation set to produce models that are robust to distribution shifts.
We then fine-tune the foundation model with the learned objective and hyperparameters and evaluate its robustness to new distribution shifts.

Our contributions are twofold.
First, we learn the fine-tuning objective itself, parameterized by weight coefficients for several different loss functions and regularizers.
This large search space gives \ours{} more granular control over adaptation.
Second, \ours{} optimizes the fine-tuning objective and hyperparameters with respect to post-adaptation performance on a proxy distribution shift: an OOD validation set. 
Importantly, this validation set is drawn from a different distribution than the final evaluation datasets and is small, containing fewer than $1\%$ of the number of examples in the fine-tuning dataset.
We illustrate the intuition behind our approach in~\cref{fig:overview} and our data assumptions in~\cref{fig:data_assumptions}.


We rigorously evaluate \ours{} on a wide array of real-world datasets and consider various types of distribution shifts, including subpopulation and domain shift.
Our experiments show that a model fine-tuned with the objective learned by \ours{} generalizes better to previously unseen distribution shifts.
With as few as $1000$ examples from a different distribution than that of the fine-tuning data, which is often readily available, \ours{} outperforms existing robust fine-tuning methods across all benchmarks. 
These gains in robustness are achieved with minimal additional compute, requiring at most $5\%$ more compute compared to standard fine-tuning.
Among other results, \ours{} achieves new state-of-the-art performance on the challenging iWildCam and FMoW benchmarks~\citep{beery2021iwildcam,koh2021wilds,christie2018functional}, outperforming the prior best methods by $6.0\%$ and $1.5\%$, respectively.



\input{sections/related.tex}
\section{Background: Hyperparameter Optimization}
\label{sec:background}
\input{sections/background.tex}

\section{\ours{}: Robust Fine-Tuning with a Learned Objective}
\label{sec:method}
\input{sections/method.tex}

\section{Experimental Setup}
\label{sec:experimental-setup}
\input{sections/experimental_setup.tex}

\section{Results}
\label{sec:results}
\input{sections/results.tex}

\subsection{Analysis}

\label{subsec:ablations}
\input{sections/ablations.tex}

\section{Conclusion}
We introduce \ours{}, a novel, data-driven approach for robust fine-tuning that learns the objective and hyperparameters.
With a small amount of data from one OOD distribution, which is often readily available, \ours{} outperforms prior fine-tuning methods.
Notably, \ours{} attains state-of-the-art performance on two WILDS benchmarks.

\textbf{Limitations.} 
\ours{} is a general recipe for robust adaptation, and our specific implementation is only one instantiation.
We expect that future work can improve upon \ours{} by investigating other loss parameterizations and meta-optimization techniques within the \ours{} framework.
While our experiments show strong results in image classification, we have not yet evaluated \ours{} in other problem settings, such as natural language processing (NLP).
We hope that our work will inspire future work on data-driven approaches for robust adaptation.

\clearpage


\section*{Acknowledgements}
We thank Kyle Hsu, Lukas Haas, and members of the IRIS lab for helpful feedback and discussions.
We also thank Sachin Goyal for help with ImageNet experiments.
This work was supported by KFAS and NSF.
A.R. is supported by Schmidt Futures, Google, Apple, and Open Philanthropy.
C.F. is supported by ONR grant N00014-21-1-2685 and the NSF CAREER award.

\bibliography{paper}
\bibliographystyle{icml2024}

\newpage
\appendix
\onecolumn
\input{sections/appendix.tex}



\end{document}

%% file: math_commands.tex
\usepackage{amsmath,amsfonts,bm}

\def\eqref#1{equation~\ref{#1}}

\def\1{\bm{1}}

\DeclareMathAlphabet{\mathsfit}{\encodingdefault}{\sfdefault}{m}{sl}
\SetMathAlphabet{\mathsfit}{bold}{\encodingdefault}{\sfdefault}{bx}{n}

\DeclareMathOperator*{\argmax}{arg\,max}
\DeclareMathOperator*{\argmin}{arg\,min}

%% file: sections/abstract.tex

Foundation models encode rich representations that can be adapted to downstream tasks by fine-tuning.
However, fine-tuning a model on one data distribution often degrades performance under distribution shifts.
Current approaches to robust fine-tuning use hand-crafted regularization techniques to constrain the fine-tuning process towards the pretrained model.
Yet, it is hard to specify how to adapt relevant characteristics of the foundation model during fine-tuning, as this depends on how the pre-training, fine-tuning, and test data distributions relate to each other.
We propose \ours{}, a data-driven approach for robust fine-tuning.
Given a task, \ours{} searches for a fine-tuning procedure that enhances out-of-distribution (OOD) generalization.
Specifically, \ours{} uses bi-level optimization to search for an objective function and hyperparameters that maximize post-adaptation performance on a small OOD validation set.
We evaluate \ours{} on nine natural distribution shifts.
Our experiments show that \ours{} significantly improves generalization to OOD inputs, outperforming existing robust fine-tuning methods. 
Notably, \ours{} achieves a new state-of-the-art on the WILDS iWildCam and FMoW benchmarks, outperforming the previous best methods by $6.0\%$ and $1.5\%$, respectively.


%% file: sections/related.tex
\section{Related Work}
\textbf{Transfer learning.}
While early research demonstrated that using features learned from pre-training on large datasets are effective for new tasks, transfer learning has evolved to optimize performance in settings with limited data ~\citep{oquab2014learning,yosinski2014transferable,sharif2014cnn}. 
Common transfer learning techniques include fine-tuning with regularization~\citep{zhang2020side,xuhong2018explicit,lee2019mixout,jiang2019smart,Li2020Rethinking,aghajanyan2020better,gouk2021distancebased,shen2021partial, KARANI2021101907} and selectively freezing pre-trained parameters~\citep{kirkpatrick2017overcoming,lee2019would,guo2019spottune,ramasesh2020anatomy,liu2021autofreeze, royer2020flexible, eastwood2021source, evci2022head2toe, eastwood2022unit, cohen2022my, touvron2022three,lee2022surgical,kumar2022sgdfinetuning}. 
However, as pre-trained models are fine-tuned for a specific distribution, their effective robustness decreases at convergence~\citep{andreassen2021evolution}.
We introduce a transfer learning framework that preserves the robustness of the pre-trained model while adapting to new tasks.

\textbf{AutoML and bi-level optimization.}
Our work leverages high-level ideas from the broader literature on meta-learning and hyperparameter optimization.
Such methods have proposed to optimize different parts of the training pipeline, including general hyperparameters \citep{hutter2011sequential,bergstra2012random,feurer2015efficient,li2017hyperband,li2018massively},
network architectures \citep{zoph2016neural,zoph2018learning,liu2018darts,real2019regularized,xu2019pc}, 
augmentation policies \citep{cubuk2019autoaugment,lim2019fast,hataya2020faster,cubuk2020randaugment}, 
optimizers \citep{bengio2013optimization,andrychowicz2016learning,wichrowska2017learned,metz2022velo,chen2023symbolic}, 
and objective functions~\citep{yu2018one,kirsch2019improving,oh2020discovering,bechtle2021meta}.
However, most of these works optimize for generalization within the training distribution and do not consider robustness to distribution shifts.
Existing works that optimize a training procedure for OOD generalization consider a structured few-shot adaptation setting \citep{li2018learning,zhang2021adaptive}, limiting their scalability to large datasets.
\citet{tian2023trainable} learn layer-specific regularization constraints to improve OOD generalization; however, they do not learn a fine-tuning objective.

\textbf{Out-of-distribution generalization.}
Maintaining good performance on data that deviates from the training distribution is crucial in many real-world applications, where models may face data from unfamiliar environments~\citep{hendrycks2019robustness,geirhos2020shortcut,gulrajani2020search,koh2021wilds}.
Numerous studies have investigated how to ensure robustness to various distribution shifts~\citep{tzeng2014deep,byrd2019effect,hendrycks2019using,arjovsky2019invariant,salman2020adversarially,liu2021just,wiles2021fine, andreassen2021evolution,creager2021environment,lee2022diversify}.
Some works have shown that despite the myriad of ways in which data distributions can change, naturally occurring distribution shifts have a surprisingly predictable effect on model performance~\citep{taori2020measuring,miller2021accuracy,baek2022agreement}, suggesting that it may be possible to \textit{learn} how to be robust to unseen natural distribution shifts.
Similar to our work, \citet{goyal2022test} meta-learn an objective, but for test-time adaptation instead of fine-tuning.
Benchmarks for OOD generalization, such as DomainBed~\citep{gulrajani2020search} and WILDS~\citep{koh2021wilds}, have discussed data assumptions for dealing with distribution shifts, and suggested that an OOD validation set can be a useful signal for hyperparameter optimization and model selection.
\ours{} leverages this signal much more directly than existing works by learning an objective for fine-tuning.
\textbf{Robust fine-tuning.}
Foundation models trained on massive datasets encode a broad range of general knowledge, enabling robust performance across various data distributions, including OOD scenarios~\citep{radford2021learning,bommasani2021opportunities}.
While in principle, foundation models should serve as a useful prior for further fine-tuning, empirical evidence shows that fine-tuning foundation models on a new task often leads to a significant drop in OOD performance~\citep{andreassen2021evolution}.
Recent works have proposed modifications to the basic fine-tuning procedure to improve OOD generalization~\citep{wortsman2022robust,kumar2022fine,wortsman2022model,goyal2022finetune,mukhoti2023fine}.
Instead of hand-designing a regularization technique, we propose a data-driven approach to \textit{learn} a more nuanced fine-tuning procedure. 
In fact, some prior works~\citep{xuhong2018explicit,lee2022surgical,goyal2022finetune} can be seen as \textit{special cases} of \ours{} since our hyperparameter search space encompasses these fine-tuning algorithms.
Our experiments in~\cref{sec:results} demonstrate that \ours{} consistently outperforms prior robust fine-tuning methods on OOD data.

%% file: sections/background.tex
We first formalize hyperparameter optimization as a means for searching the space of fine-tuning procedures in \cref{sec:method}.
Hyperparameters are predefined properties of the learning algorithm which are not learned during training, such as network architecture, learning rate, and regularization strength.
A good choice of hyperparameters can significantly improve model performance, yet optimal hyperparameters vary, depending on the data distribution and evaluation metrics.

Formally, we denote the learning algorithm as $\LA$ and its hyperparameters as $\phi \in \Phi$, where $\Phi$ is the hyperparameter space.
We denote the training and validation datasets as $\Dtr$ and $\Dval$, respectively.
These datasets are disjoint and typically drawn from the same distribution.
The learning algorithm $\LA$ produces a model by training on $\Dtr$ with hyperparameters $\phi$.
We denote the resulting model as $\LA(\phi, \Dtr)$.
Hyperparameter optimization finds hyperparameters that maximize some performance metric $\mathrm{Perf}(f, \Dval)$ which depends on the model $f$ and the validation dataset $\Dval$.
Examples of performance metrics include top-1 accuracy, macro F1 score, and worst-region accuracy.
We formalize hyperparameter optimization as
\begin{equation}
    \label{eq:hyperopt}
    \phi^* = \argmax_{\phi \in \Phi} \mathbb{E} \big[ \overbrace{\mathrm{Perf}(\underbrace{\LA(\phi, \Dtr)}_{\text{Learned Parameters}}, \Dval)}^{\text{Validation Set Performance}} \big].
\end{equation}
Here, the expectation is taken over any randomness in the learning algorithm $\LA$, such as input data shuffling or random initialization.
The optimized hyperparameters $\phi^*$ are subsequently used to train the model.

Hyperparameter optimization methods typically start with randomly initialized model parameters and use a validation set $\Dval$, drawn from the same distribution as the training data, to adjust hyperparameters.
The problem of robust fine-tuning, however, begins with pre-trained model parameters and aims to achieve high performance on test data that diverges from the fine-tuning distribution.
In the next section, we describe how we leverage hyperparameter optimization to learn a robust adaptation procedure.


%% file: sections/method.tex

In this section, we present \ours{}, a data-driven approach for robust fine-tuning.
Given a foundation model and task-specific data, \ours{} uses bi-level optimization to learn a fine-tuning objective and optimal hyperparameters.
There are two key components in \ours{}.
First, learning the fine-tuning objective enables a more precise adaptation to task-specific data.
Second, \ours{} optimizes the objective and hyperparameters for performance under a proxy distribution shift, which enhances OOD generalization.

\subsection{Problem Setting: Robust Fine-Tuning}

\textbf{Data assumptions.} We assume access to two datasets:
(1) a large fine-tuning dataset $\Dtr$ from the training distribution $\Ptr$, and 
(2) a small held-out OOD validation set $\Dval$ from a different distribution $\Pval$.
The validation set $\Dval$ is much smaller than $\Dtr$, containing as few as $1000$ examples, and is only used for outer-level optimization, not for fine-tuning.
We then fine-tune on $\Dtr$ with the found hyperparameters and test the fine-tuned model on several OOD distributions $\Pood^1, \Pood^2, \dots$.
Note that these test distributions are distinct from both $\Ptr$ and $\Pval$, and are never shown to the model before final evaluation.
Following prior works on robust fine-tuning, all distributions are assumed to be related to the same task.
We consider domain and subpopulation shifts arising from differences in data collection. 
For instance, in animal recognition from camera trap images, each distribution may represent different locations \citep{beery2021iwildcam,koh2021wilds}.
Such data is often readily available, as shown in the WILDS and DomainBed benchmarks \citep{koh2021wilds,gulrajani2020search}, and can be constructed by partitioning by time or geographical location.

\textbf{Notation and concrete setup.} 
Let $f$ denote a pre-trained foundation model with parameters $\theta$, which we adapt by fine-tuning on $\Dtr$.
We denote the fine-tuning algorithm as $\LA(\phi, \Dtr)$, where $\phi$ specifies both the fine-tuning objective and other hyperparameters such as learning rate.
In our experiments, the performance metric $\mathrm{Perf}$ follows the standards in each evaluation setting and is either top-1 accuracy, worst-region accuracy, or macro F1.


\input{tables/alg.tex}

\subsection{Learning a Fine-Tuning Objective via Bi-Level Optimization}




\textbf{Formulation.}
We frame the problem of learning a fine-tuning objective as a bi-level optimization problem with the form~\cref{eq:hyperopt}, which we can solve using existing black-box hyperparameter optimizers.
Practically, solving this bi-level optimization problem involves fine-tuning the same foundation model multiple times, each time with different ``hyperparameters'' $\phi^1, \phi^2, \ldots$ to obtain different final fine-tuned models $f^1=\LA(\phi^1, \Dtr), f^2=\cdots$. 
We then evaluate the quality of each $\phi^i$ through the performance of $f^i$ on the OOD validation set; by aggregating the information from these different fine-tuning runs, we can fine good hyperparameters $\phi^*$.
At test time, we fine-tune the initial foundation model with $\phi^*$ and evaluate its performance on novel OOD distributions to assess its generalization capabilities.
We summarize this procedure in Algorithm~\ref{alg:hyperopt}.

\textbf{Parameterization of the fine-tuning objective.} 
The main novelty of \ours{} is that we parameterize the fine-tuning objective, which allows outer-loop optimization to search over an expressive space of fine-tuning algorithms.
This extra expressivity is important: for robust fine-tuning, the algorithm must selectively \textit{ignore} some aspects of the training distribution in order to better generalize to novel OOD distributions.
Specifically, we allow $\Phi$ to express the fine-tuning objective by learning weights for a pre-defined set of loss functions and regularizers.
Note that the cross-entropy and hinge losses similarly guide the model's predictions towards the correct label, but with a different amount of penalty for severely incorrect samples.
As another example, the contrastive loss on image-text pairs~\citep{radford2021learning,goyal2022finetune} also guides the model's predictions towards the correct label, but in a way that incorporates language embeddings, which is entirely different from how the cross-entropy and hinge losses operate.

Therefore, we consider weight coefficients for nine different loss functions and regularizers: cross-entropy loss, hinge loss, image-text contrastive loss, entropy, confidence minimization, L1 norm, L2 norm, L1 and L2 distance to pretrained model parameters.
We select these losses to address various aspects of model learning and generalization, such as classification accuracy, decision confidence, and prevention of overfitting, ensuring a nuanced and effective model adaptation.
We denote these loss weight coefficients as $W = \{w_1, w_2, \dots, w_9\}$.
Denoting the $i$-th loss function or regularizer as $\mathcal{L}_i$, the total loss $\mathcal{L}$ is the weighted sum $\mathcal{L} = \sum_{i=1}^{9}w_i \mathcal{L}_i$.
To enable finer control over the speed and stability of adaptation during fine-tuning, we additionally learn hyperparameters for the fine-tuning optimizer: learning rate $\eta$, weight decay $\delta$, and random seed $\sigma$.
Our complete set of hyperparameters for outer-loop optimization is thus $\phi=(W, \eta, \delta, \sigma)$.

\textbf{Bi-level optimization algorithm.} 
We need a black-box optimizer that can effectively search through the high-dimensional hyperparameter space $\Phi$.
We use the Tree-structured Parzen Estimator~\citep[TPE]{NIPS2011_86e8f7ab} for this purpose, as its design allows for efficient processing in high-dimensional spaces and yields effective results at the order of hundreds of evaluations.
TPE creates a probabilistic model from previous hyperparameter evaluations, and differentiates between better and worse hyperparameter configurations by comparing their likelihood under two separate models.
We use the TPE implementation from the \texttt{optuna} library~\citep{akiba2019optuna}. 
Experiments in~\cref{tab:hpo_algorithm} demonstrate that in the context of learning a fine-tuning objective, TPE outperforms other bi-level optimizers such as random search, quasi Monte Carlo, and Bayesian optimization.

As the prior distribution for TPE, we use an appropriately scaled distribution for each hyperparameter, operating in log-space for the weight coefficients and learning rate since the optimal values can vary over several orders of magnitude:
\begin{align*}
  w_i \sim \text{LogUniform}(w_{\text{min}}, w_{\text{max}}), \quad &\rho \sim [0, 100]  \\
  \eta \sim \text{LogUniform}(\eta_{\text{min}}, \eta_{\text{max}}), \quad &\delta \sim \text{Uniform}(0.0, 1.0). 
\end{align*}
We find that in practice, $(w_{\text{min}}, w_{\text{max}}) = (10^{-4},10)$ is an effective range for all $w_i$.
Given a model's conventional learning rate $\eta^*$, we set $(\eta_{\text{min}}, \eta_{\text{max}}) = (10^{-2} \eta^*, 10^2 \eta^*)$.

\textbf{Intuition for OOD evaluation.} 
Optimizing the objective with respect to performance on a small out-of-distribution (OOD) validation set $\Dval$ can guide the model to learn more generalizable parameters during fine-tuning.
We conduct a didactic toy experiment in~\cref{sec:didactic-experiment}, which illustrates our key intuition: a small validation set reveals \textit{how} the data distribution changes under natural shifts, informing how best to adapt a given foundation model.
Our experiments in \cref{sec:results} confirm that \ours{} consistently improves generalization to novel OOD distributions, outperforming state-of-the-art methods.

\textbf{Computational cost.} 
The bi-level optimization of \ours{} introduces minimal computational overhead, requiring at most $5\%$ additional compute compared to one standard fine-tuning run on most benchmarks.
Each evaluation of the objective and hyperparameters requires a small number of inner loop gradient steps on the fine-tuning dataset.
\ours{} is therefore computationally inexpensive, and highly practical for even large fine-tuning problem settings.
Hyper-hyperparameters are detailed in \cref{sec:appendix-hyperparams}.

%% file: tables/alg.tex
\newcommand{\algcomment}[1]{%
\hfill {\raggedleft\textcolor{gray!60}{// #1}\par}
}
\begin{algorithm}[t]
\caption{\label{alg:hyperopt} \ours{}}
\begin{algorithmic}[1]
    \REQUIRE Hyperparameter Optimizer (HPO)
    \REQUIRE ID Training Data $\Dtr$, OOD Validation Data $\Dval$
    \FOR{$\phi \leftarrow \text{HPO.Sample}()$}
        \STATE $f_{ft} \leftarrow \LA(\Dtr, \mathcal{L}_\phi)$ \algcomment{Short fine-tuning run with sampled loss}
        \STATE $p \leftarrow \mathrm{Perf}(f_{ft}, \Dval)$ \algcomment{Evaluation on OOD Val set}
    \ENDFOR
    \STATE $\phi^* \leftarrow \mathrm{HPO}.\mathrm{Best}()$ \algcomment{Get best hyperparameters}
    \STATE $f^* \leftarrow \LA(\Dtr, \mathcal{L}_{\phi^*})$ \algcomment{Fine-tune with loss $\mathcal{L}_{\phi^*}$}
    \STATE \textbf{Return} $f^*$
\end{algorithmic}
\end{algorithm}

%% file: sections/experimental_setup.tex
\textbf{Evaluation.} We fine-tune pre-trained CLIP models from the \texttt{open-clip} repository~\citep{radford2021learning,ilharco_gabriel_2021_5143773}. 
We use the CLIP \texttt{ViT-B/16} model from OpenAI as our default model, unless specified otherwise.
Our SoTA results on WILDS-iWildCam and WILDS-FMoW use the CLIP \texttt{ViT-L/14-336px} model.
We use text templates used in prior work~\citep{radford2021learning,wortsman2022robust} to generate zero-shot classification weights.

Following prior works on robust fine-tuning, we evaluate \ours{} on nine \textit{natural distribution shifts}, few-shot learning, and transfer learning settings \cite{taori2020measuring, radford2021learning, wortsman2022robust, kumar2022fine, goyal2022finetune}.
These include five ImageNet distribution shifts (IN-V2, IN-R, IN-Sketch, ObjectNet, and IN-A) \citep{recht2019imagenet, hendrycks2021many, wang2019learning, barbu2019objectnet, hendrycks2021natural}, WILDS-iWildCam \citep{koh2021wilds,beery2021iwildcam}, WILDS-FMoW \citep{koh2021wilds,christie2018functional}, CIFAR-10.1 \citep{recht2018cifar10.1, torralba2008tinyimages}, and CIFAR-10.2 \citep{lu2020harder}. 
We use up to $1000$ examples from ImageNet-C, OOD validation splits of WILDS-FMoW and WILDS-iWildCam, and CIFAR-10-C as the OOD validation sets.
We also evaluate on few-shot classification (\cref{tab:fewshot-binary-classification} and \cref{fig:imagenet-fewshot-effective-robustness-plots}) and transfer learning (\cref{tab:transfer_learning}).


\begin{table}[b]
\vspace{-3mm}
\centering
\begin{tabular}{lccc}
\toprule
\textbf{Dataset} & \textbf{Steps} & \textbf{Trials} & \textbf{$N_\textrm{val}$} \\
\midrule
iWildCam & 10 & 500 & 1000 \\
FMoW & 10 & 500 & 1000 \\
CIFAR-10 & 10 & 100 & 100 \\
Flowers & 50 & 500 & 500 \\
Cars & 50 & 500 & 1000 \\
ImageNet & 100 & 500 & 1000  \\
\bottomrule
\end{tabular}
\caption{\label{tab:hyperhyper}
\ours{} training settings for each dataset.
}
\end{table}

\textbf{Effective robustness and weight ensembling curves.} 
We primarily evaluate robustness to distribution shifts using average OOD performance.
We also compare methods with weight ensembling of the zero-shot and fine-tuned models \citep{wortsman2022robust}, which enhances ID and OOD performance in a way orthogonal to fine-tuning.
We interpolate model weights with ten mixing coefficients, \( \alpha \), and report results using the coefficient that maximizes ID validation accuracy.
Finally, we plot ID and OOD performance across these different interpolation coefficients in \cref{fig:effective-robustness-plots} to visualize \textit{effective robustness} -- gains in accuracy beyond the zero-shot model, which corresponds to \textit{vertical distance} between curves.

\textbf{Baselines.} 
We compare \ours{} against several methods for adapting pretrained models. 
We include two standard transfer learning methods that minimize cross-entropy loss: linear probing (LP) and full fine-tuning (FT).
We also compare with recent works in robust fine-tuning: 
L2-SP~\citep{li2018massively}, which fine-tunes with an L2 regularization term towards pretrained weights; 
LP-FT~\citep{kumar2022fine}, which performs linear probing followed by full fine-tuning; 
Freeze-Embed~\citep{kumar2022sgdfinetuning}, which freezes the embedding layer during fine-tuning;
and FLYP~\citep{goyal2022finetune}, which fine-tunes with the CLIP pretraining loss -- a contrastive loss between image embeddings and class-descriptive prompt embeddings.
As we are interested in OOD performance, we also compare against OOD generalization methods, including Group DRO~\citep{sagawa2019distributionally}, ABSGD~\citep{qi2020attentional}, LISA~\citep{yao2022improving}, DFR~\citep{kirichenko2022last}, and Copy-Paste~\citep{gao2023out}.
We additionally evaluate all methods with weight ensembling (WiSE-FT)~\citep{wortsman2022robust}, which is shown to improve OOD performance in an orthogonal way to other robust fine-tuning methods.

\textbf{Bi-level optimization details.} 
We summarize the number of inner-loop gradient steps, outer-loop fine-tuning trials, and validation set size for each dataset in \cref{tab:hyperhyper}.
All of these values were selected based on performance on an ID validation set.
We provide additional details on our training protocol in~\cref{sec:appendix-hyperparams}.

%% file: sections/results.tex
\input{tables/iwildcam_sota.tex}

In this section, we present the main experimental findings for \ours{}. 
We show that \ours{} improves OOD generalization on nine distribution shifts.
We present additional results in the low-data and IID transfer learning regimes in \cref{sec:few-shot} and \cref{sec:transfer-learning}.
These findings highlight the effectiveness of \ours{} for robust fine-tuning in a variety of settings.
Finally, we investigate several aspects of the \ours{} method, including transferability of the learned objective across datasets, the choice of validation dataset and hyperparameter optimization algorithm.
Further experimental details are in \cref{sec:exp_details}.

\input{tables/fmow_sota.tex}

\begin{figure*}[t]
\vspace{-1mm}
\centering 
\includegraphics[width=1.0\linewidth]{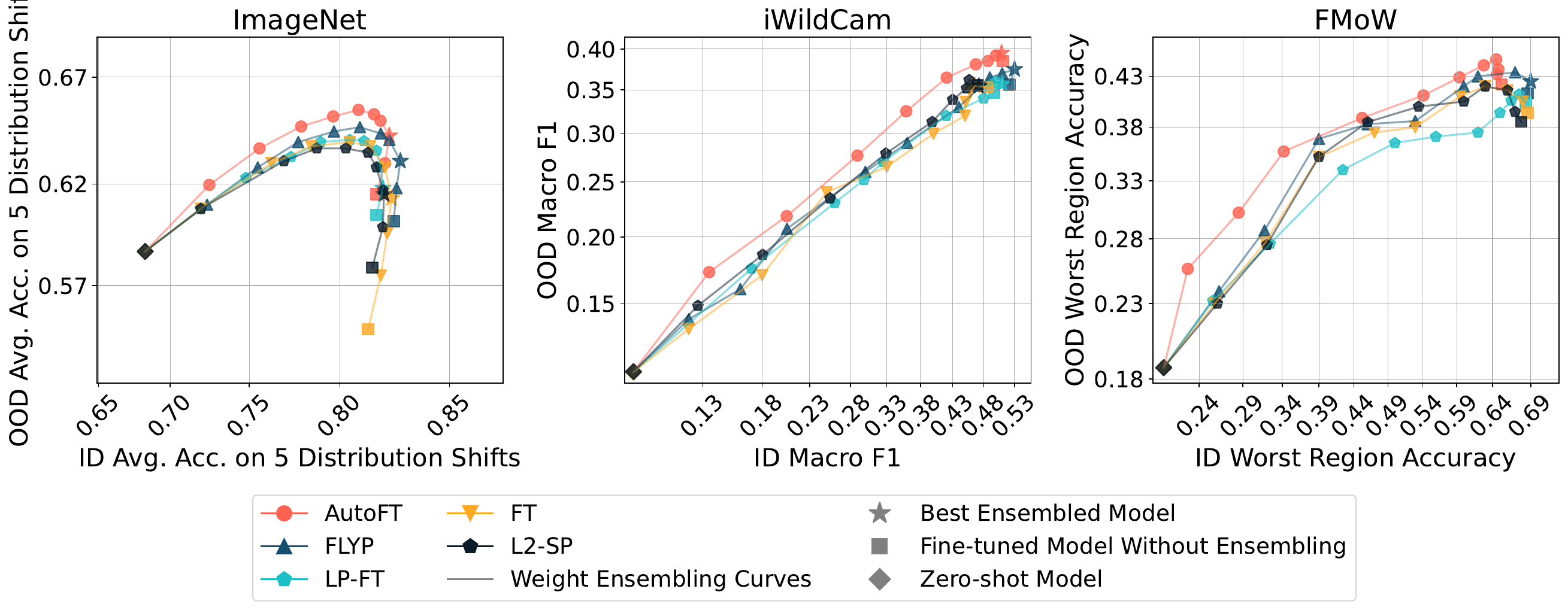}
\vspace{-5mm}
\caption{\ours{} outperforms existing methods, both with and without weight ensembling~\citep{wortsman2022robust}.
Here, we show the ID-OOD performance curves obtained by linearly interpolating weights of the fine-tuned model weights with the zero-shot model.
}
\label{fig:effective-robustness-plots}
\end{figure*}

\begin{table*}[t]
\vspace{0mm}
    \centering
    \input{tables/big_table_arxiv}
\end{table*}

\input{tables/hparam_transfer.tex}
\input{tables/validation_sets.tex}


\subsection{Distribution Shifts}

\textbf{Improvements in OOD generalization.} 
We evalute \ours{} on three common robust fine-tuning benchmarks: ImageNet, WILDS-FMoW, and WILDS-iWildCam. 
This evaluation includes diverse distribution shift conditions including subpopulation and domain shifts.
ID and OOD performance reported in~\cref{tab:big_distribution_shift_table} show that \ours{} consistently outperforms existing methods for robust fine-tuning in OOD metrics.
Furthermore, these gains are maintained when ensembling initial and fine-tuned models, suggesting that the benefits from \ours{} are complementary to that of ensembling~\citep{wortsman2022robust,wortsman2022model}.
We show additional experiments on CIFAR-10 and related distribution shifts in~\cref{tab:cifar10}, which show the same conclusion.

\textbf{State-of-the-art performance on iWildCam and FMoW.}
To further assess whether the robustness gains from \ours{} scale to larger foundation models, we evaluate \ours{} with a \texttt{ViT-L/14@336px} model, following the largest scale experiment in \citet{goyal2022finetune}.
Results in \cref{tab:sota_iwild} show that \ours{} improves upon the prior state-of-the-art method, FLYP \citep{goyal2022finetune}, by $6.0\%$ in OOD macro-F1.
Furthermore, on WILDS-FMoW~\cref{tab:sota_fmow}, \ours{} outperforms a Transformer-specific parameter freezing method~\citep{kumar2022sgdfinetuning}, the prior state-of-the-art, by $1.5\%$ in OOD worst-region accuracy.
Additionally, \ours{} outperforms the compute-intensive Model Soups~\citep{wortsman2022model} on both benchmarks; this method ensembles more than $70$ models fine-tuned with different augmentations and hyperparameters.

\subsection{Few-Shot Classification}
\label{sec:few-shot}

Few-shot classification serves as an important benchmark in applications where there are small quantities of labeled data available for fine-tuning.
Few-shot binary classification is particularly challenging, given the small number of training examples.
We evaluate on 4, 16, and 32 shot binary classification tasks from the PatchCamelyon and Rendered-SST2 datasets, following \citet{radford2021learning} and \citet{goyal2022finetune}.
PatchCamelyon contains digital pathology images for the detection of metastatic tissue.
Rendered-SST2 focuses on optical character recognition for classifying text sentiment.

\cref{tab:fewshot-binary-classification} shows that \ours{} still improves effective robustness with limited data, outperforming FLYP by $3.7\%$ and full fine-tuning by $3.9\%$ in a challenging 4-shot classification task on Rendered-SST2.

\begin{figure*}
\vspace{-1mm}
\begin{minipage}[b]{0.72\textwidth}
\centering
\resizebox{0.99\linewidth}{!}{
\begin{tabular}{lcccccc}
\toprule
& \multicolumn{3}{c}{PatchCamelyon} & \multicolumn{3}{c}{SST2} \\
\cmidrule(lr){2-4} \cmidrule(lr){5-7}
$k$ (shots) & 4 & 16 & 32 & 4 & 16 & 32 \\
\midrule
Zeroshot & 56.5 (-) & 56.5 (-) & 56.5 (-) & 60.5 (-) & 60.5 (-) & 60.5 (-) \\
LP & 60.4 (4.0) & 64.4 (3.7) & 67.0 (4.4) & 60.8 (1.8) & 61.9 (1.4) & 62.9 (1.3) \\
FT & 63.1 (5.5) & 71.6 (4.6) & 75.2 (3.7) & 61.1 (0.7) & 62.4 (1.6) & 63.4 (1.9) \\
LP-FT & 62.7 (5.3) & 69.8 (5.3) & 73.9 (4.6) & 60.9 (2.4) & 62.9 (1.9) & 63.6 (1.4) \\
FLYP & 66.9 (5.0) & 74.5 (2.0) & 76.4 (2.4) & 61.3 (2.7) & 65.6 (2.1) & 68.0 (1.7) \\
\midrule
\oursbf{} & \textbf{68.1 (5.1)} & \textbf{76.8 (2.9)} & \textbf{79.5 (2.0)} & \textbf{65.0 (3.8)} & \textbf{67.5 (1.1)} & \textbf{69.0 (1.1)} \\
\bottomrule
\end{tabular}
}
\vspace{-2mm}
\caption{In binary few-shot classification, \ours{} outperforms existing robust fine-tuning methods. 
\ours{} outperforms FLYP by $3.1\%$ and full fine-tuning by $4.3\%$ in 32-shot classification on PatchCamelyon.
}
\label{tab:fewshot-binary-classification}
\end{minipage}
\hfill
\begin{minipage}[b]{0.26\textwidth}
\centering
\begin{tabular}{lcc}
\toprule
Examples & ID & OOD \\
\midrule
200          & 48.9    & 37.2     \\
1000         & \textbf{51.0} & \textbf{38.3} \\
5000         & 50.4    & 38.0     \\
10000        & 48.9    & 37.8     \\
\bottomrule
\end{tabular}
\caption{
\label{tab:val-set-size}
Test performance of \ours{} on WILDS-iWildCam with varying validation set sizes. 
Increasing size beyond $1000$ shows minimal additional benefits.}
\end{minipage}
\end{figure*}




\subsection{Transfer Learning}
\label{sec:transfer-learning}

We additionally evaluate \ours{} on standard transfer learning tasks, where the evaluation data is drawn from the same distribution as both the training and validation data.
We compare \ours{} against other fine-tuning methods on several transfer learning datasets: CalTech-101, Stanford Cars, and Flowers-102.
Results in~\cref{tab:transfer_learning} demonstrate that \ours{} is competitive with these baselines, but the advantages from learning an objective are less pronounced compared to previous experiments.
This suggests that problem-specific learned fine-tuning objectives are most helpful in settings with high underspecification, such as OOD evaluation conditions or few-shot learning, confirming our initial intuition in~\cref{sec:method}.


%% file: tables/iwildcam_sota.tex

\begin{table}[t]
\centering
\begin{tabular}{@{}llcc@{}}
\toprule
Method & Architecture & ID & OOD \\
\midrule
Group DRO & ResNet50 & 37.5 (1.9) & 23.8 (2.0) \\
ABSGD & ResNet50 & 47.5 (1.6) & 33.0 (0.6) \\
Copy-Paste & ResNet50 & 50.2 (1.6) & 36.5 (0.9) \\
ERM & PNASNet & 52.8 (1.4) & 38.5 (0.6) \\
ERM & ViTL & 55.8 (1.9) & 41.4 (0.5) \\
Model Soups & ViTL & 57.6 (1.9) & 43.3 (1.0) \\
FLYP & ViTL-336px & 59.9 (0.7) & 46.0 (1.3) \\
\midrule
\oursbf{} & ViTL-336px & 63.5 (0.5) & \textbf{52.0 (0.4)} \\
\bottomrule
\end{tabular}
\caption{
    \textbf{iWildCam SoTA results.}
    \ours{} with weight ensembling attains state-of-the-art OOD performance on the WILDS-iWildCam benchmark with a \texttt{ViT-L/14-336px} backbone, surpassing all prior entries on the WILDS leaderboard \citep{koh2021wilds}.}
\label{tab:sota_iwild}
\vspace{-3mm}
\end{table}

%% file: tables/fmow_sota.tex
\begin{table}[t]
\centering
\begin{tabular}{@{}llcc@{}}
\toprule
Method & Architecture & ID & OOD \\
\midrule
Group DRO & DenseNet121 & 51.2 (0.4) & 31.1 (1.7) \\
LISA & DenseNet121 & 52.8 (1.2) & 35.5 (0.8) \\
ERM w/ aug & DenseNet121 & 55.5 (0.4) & 35.7 (0.3) \\
DFR & DenseNet121 & 53.4 (0.4) & 42.8 (0.4) \\
ERM & ViTL & 66.9 (0.2) & 46.1 (0.6) \\
Model Soups & ViTL & 69.5 (0.1) & 47.6 (0.3) \\
Freeze-Embed & ViTL-336px & 68.3 (0.4) & 50.3 (1.1) \\
\midrule
\oursbf{} & ViTL-336px & 72.1 (0.1) & \textbf{51.8 (0.4)} \\
\bottomrule
\end{tabular}
\caption{
    \textbf{FMoW SoTA results.}
    \ours{} with weight ensembling attains state-of-the-art OOD performance on the WILDS-FMoW benchmark with a \texttt{ViT-L/14-336px} backbone, surpassing all prior entries on the WILDS leaderboard \citep{koh2021wilds}.}
\label{tab:sota_fmow}
\vspace{-3mm}
\end{table}

%% file: tables/big_table_arxiv.tex
\centering
\resizebox{\textwidth}{!}{
\begin{tabular}{@{}ccccc|cccc|cc@{}}
\toprule
& \multicolumn{4}{c|}{ImageNet} & \multicolumn{4}{c|}{iWILDCam} & \multicolumn{2}{c}{FMoW} \\
\cmidrule(lr){2-5} \cmidrule(lr){6-9} \cmidrule(lr){10-11}
& \multicolumn{2}{c}{Without Ensembling} & \multicolumn{2}{c|}{With Ensembling} & \multicolumn{2}{c}{Without Ensembling} & \multicolumn{2}{c|}{With Ensembling} & \multicolumn{2}{c}{Without Ensembling} \\
\cmidrule(lr){2-3} \cmidrule(lr){4-5} \cmidrule(lr){6-7} \cmidrule(lr){8-9} \cmidrule(lr){10-11}
Methods & ID & OOD & ID & OOD & ID & OOD & ID & OOD & ID & OOD \\
\midrule
Zeroshot & 68.3 (-) & 58.7 (-) & 68.3 (-) & 58.7 (-) & 8.7 (-) & 11.0 (-) & 8.7 (-) & 11.0 (-) & 20.4 (-) & 18.7 (-) \\
LP & 79.9 (0.0) & 57.2 (0.0) & 80.0 (0.0) & 58.3 (0.0) & 44.5 (0.6) & 31.1 (0.4) & 45.5 (0.6) & 31.7 (0.4) & 48.2 (0.1) & 30.5 (0.3) \\
FT & 81.4 (0.1) & 54.8 (0.1) & 82.5 (0.1) & 61.3 (0.1) & 48.1 (0.5) & 35.0 (0.5) & 48.1 (0.5) & 35.0 (0.5) & 68.5 (0.1) & 39.2 (0.7) \\
L2-SP & 81.6 (0.1) & 57.9 (0.1) & 82.2 (0.1) & 58.9 (0.1) & 48.6 (0.4) & 35.3 (0.3) & 48.6 (0.4) & 35.3 (0.3) & \textbf{68.6 (0.1)} & 39.4 (0.6) \\
LP-FT & 81.8 (0.1) & 60.5 (0.1) & 82.1 (0.1) & 61.8 (0.1) & 49.7 (0.5) & 34.7 (0.4) & 50.2 (0.5) & 35.7 (0.4) & 68.4 (0.2) & 40.4 (1.0) \\
FLYP & 82.6 (0.0) & 60.2 (0.1) & 82.9 (0.0) & 63.2 (0.1) & 52.2 (0.6) & 35.6 (1.2) & 52.5 (0.6) & 37.1 (1.2) & \textbf{68.6 (0.2)} & 41.3 (0.8) \\
\midrule 
\ours{} & 81.8 (0.1) & \textbf{61.5 (0.1)} & 82.4 (0.1) & \textbf{64.3 (0.1)} & 51.0 (0.5) & \textbf{38.3 (0.5)} & 51.3 (0.5)  & \textbf{39.3 (0.5)} & 67.1 (0.3) & \textbf{42.3 (0.5)} \\
\bottomrule
\end{tabular}
}
\caption{
\ours{} outperforms all baselines, both with and without ensembling.
Without ensembling, \ours{} improves OOD performance by $1.3\%$ on ImageNet, $2.7\%$ on WILDS-iWildCam, and $1.0\%$ on WILDS-FMoW. 
These improvements are preserved with weight ensembling.
}
\label{tab:big_distribution_shift_table}

%% file: tables/hparam_transfer.tex

\begin{table}[t!]
\centering
\resizebox{\columnwidth}{!}{
\begin{tabular}{l|cc}
\toprule
Method & ID & OOD \\ 
\midrule
FT & 48.1 & 35.0 \\ 
\ours{}, transferred objective & 49.9 & 35.5 \\
\ours{} & \textbf{51.0} & \textbf{38.3} \\
\bottomrule
\end{tabular}
}
\vspace{-2mm}
\caption{
Learning a fine-tuning objective on the task of interest (\ours{}) outperforms fine-tuning with an objective learned on a different task (\ours{}, transferred objective).
This suggests that the learned objective is not universal, but specific to the task on which it is optimized.
}
\label{tab:hparam-transfer}
\end{table}

%% file: tables/validation_sets.tex
\begin{table}[t!]
\centering
    \begin{tabular}{ll|cc}
    \toprule
    Method & Validation Set & OOD Test \\
    \midrule
    FT & CIFAR-10 & 93.6 (0.2) \\
    \midrule
    \ours{} & CIFAR-10 & 95.1 (0.2) \\
    & CIFAR-10-C & \textbf{95.5 (0.2)} \\
    & CINIC & 94.5 (0.2) \\
    & CIFAR-10.1 & 94.8 (0.3) \\
    & CIFAR-10.2 & 95.3 (0.3) \\
    \bottomrule
    \end{tabular}
\vspace{-2mm}
    \caption{
    \label{tab:varying_validation_sets}
        \ours{} outperforms fine-tuning with several different validation sets.
        We fine-tune on CIFAR-10 and use $100$ examples from either the ID distribution (CIFAR-10) or a non-ID distribution (CIFAR-10-C, CINIC, CIFAR-10.1, CIFAR-10.2) to learn the objective and hyperparameters.
        Evaluation is on held-out examples from CIFAR-10.1 and CIFAR-10.2.
    }
\end{table}


%% file: sections/ablations.tex
\textbf{Transferability of the learned objective.} 
We evaluate models fine-tuned on WILDS-iWildCam with an objective learned on the same WILDS-iWildCam task (\ours{}) or a different task, WILDS-FMoW (\ours{}, transferred objective).
We additionally compare with standard fine-tuning (FT) in \cref{tab:hparam-transfer}.
Fine-tuning with an objective learned on a different task (\ours{}, transferred objective) degrades performance, suggesting that the learned objective is not universal, but tailored to the task on which it is optimized.


\textbf{Choice of dataset for evaluating the learned hyperparameters.} 
\begin{table} \centering
\resizebox{0.8\columnwidth}{!}{
\begin{tabular}{l|cc}
\toprule
Validation Set & ID Test & OOD Test \\
\midrule
ID Validation & \textbf{52.3}    & 35.8     \\
OOD Validation & 51.0    & \textbf{38.3}     \\
\bottomrule
\end{tabular}
}
\caption{
\label{tab:val-set-choice}
Using a non-ID validation set improves OOD generalization compared to using an ID validation set. 
We evaluate \ours{} on WILDS-iWildCam, using either the ID or OOD validation split from the WILDS benchmark to learn the objective and hyperparameters.
Using an ID validation set improves ID performance, while using a non-ID validation set improves OOD performance.
}
\end{table}
In \cref{tab:val-set-choice}, we run \ours{} with different validation sets -- the official ID validation, OOD validation, and OOD test splits from the WILDS-iWildCam benchmark. Optimizing hyperparameters with respect to an ID validation set improves ID performance, while using a non-ID validation set  improves OOD generalization.

\textbf{Choice of hyperparameter optimization algorithm.} 
\begin{table} \centering
\begin{tabular}{lcc}
\toprule
Hyperparameter Optimization & ID Test & OOD Test \\
\midrule
Random                           & 29.8    & 24.5     \\
Quasi Monte Carlo                & 48.2    & 34.7     \\
Bayesian Optimization            & 49.9    & 36.5     \\
Tree-Structured Parzen Estimator & \textbf{51.0} & \textbf{38.3}     \\
\bottomrule
\end{tabular}
\caption{\label{tab:hpo_algorithm}
We run \ours{} with various hyperparameter optimization algorithms given the same computational budget.
Evaluation is on WILDS-iWildCam.
The Tree-Structured Parzen Estimator (TPE) outperforms other methods by a large margin.
}
\end{table}
We assess the effect of different hyperparameter optimization algorithms on the performance of \ours{}.
We run \ours{} with three different hyperparameter optimization methods: random search, quasi Monte Carlo, and Bayesian optimization.
Results in \cref{tab:hpo_algorithm} show that Tree-Structured Parzen Estimator (TPE) outperforms other hyperparameter optimization methods.

\textbf{Effect of validation set size.}
We examine the effect of validation set size on performance in \cref{tab:val-set-size}. 
Across all validation set sizes, \ours{} yields improvements in OOD performance compared to the leading fine-tuning baseline (FLYP: $35.6\%$).
Given a fixed number of inner steps and hyperparameter evaluations, there's an optimal range for validation set size, beyond which additional benefits are minimal.


%% file: sections/appendix.tex
\section{Intuition and Didactic Experiment for \ours{}}
\begin{figure*}[t!]
  \centering 
  \includegraphics[height=38mm]{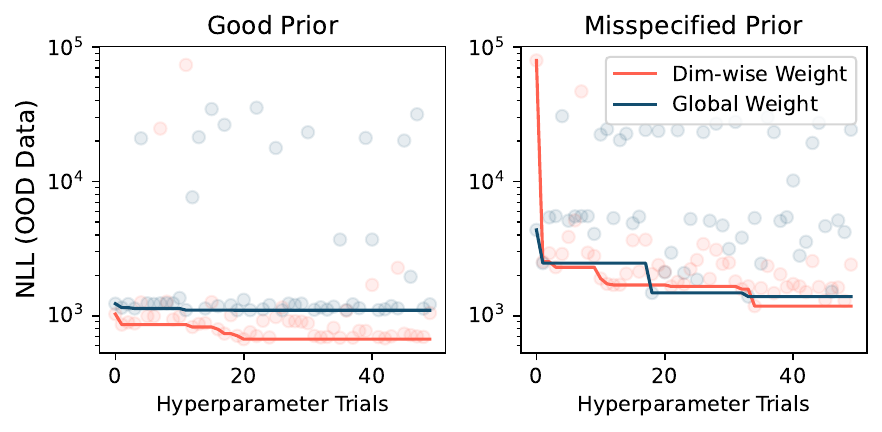} \hfill
  \includegraphics[height=38mm]{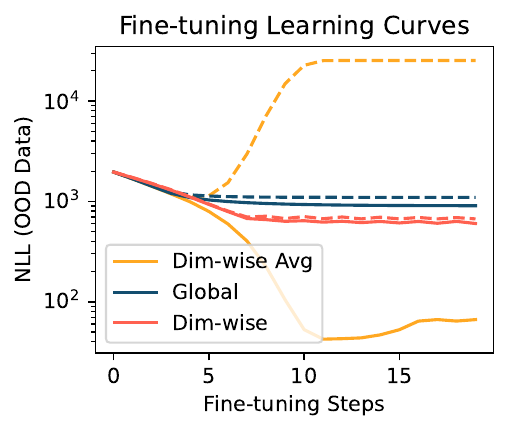} \hfill
  \includegraphics[height=38mm]{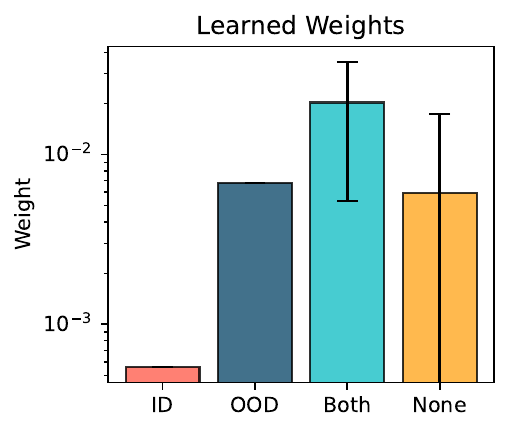}
\caption{ \label{fig:toy}
\textbf{Didactic experiment.} 
\textit{Left two}: learning curves for fitting a Gaussian distribution to toy data.
Hyperparameter optimization with a larger hyperparameter space (red) leads to better OOD performance only when the prior is informative.
\textit{Right}: fine-tuning learning curves with learned hyperparameters.
Averaging the dimension-wise weights (yellow) results in massively overfitting to the ID data (solid) and underperforming OOD (dashed).
The dimension-wise hyperparameters (red) show the best generalization.
\textit{Far right}: visualization of learned loss weights $w_i$ for each dimension. 
The learned weights are lowest for the dimensions where ID alone differs from the prior (red), indicating that the model learns to ignore the ID data in these dimensions.
}
\end{figure*}

\label{sec:didactic-experiment}

To illustrate how hyperparameter optimization can lead to more effective adaptation of a partially useful prior, we consider a simple experiment.
We focus on synthetic datasets of vectors drawn from zero-mean Gaussian distributions with different variances.

We assume that the prior unit variance distribution is helpful but that the data exhibits deviation from this prior along a few dimensions. 
The prior distribution is a unit Gaussian, i.e., it has variance \([\highvar, \highvar, \ldots]\).
Training data $\Dtr$ is drawn from a distribution with variance \([\lowvar, \lowvar, \lowvar, \lowvar, \highvar, \highvar, \ldots]\), while the validation data $\Dval$ is drawn from a distribution with variance \([\lowvar, \lowvar, \lowvar, \highvar, \lowvar, \highvar, \ldots]\).
We fit the parameters of a Gaussian distribution $q_\theta$ to the training data using a variational Bayes objective by minimizing the following loss:
\begin{equation}
  \argmin_\theta \kld{q_\theta}{p} + \sum_{i=1}^{10} w_i \text{NLL}_i(q_\theta, \mathcal{D}).
\end{equation}

We consider ``fine-tuning'' as starting from the prior distribution. 
In this toy model, dimension-wise weight hyperparameters $w_i$ dictate the degree of influence of the prior versus the training data on each dimension, analogous to how the fine-grained hyperparameters in \ours{} balance the foundation model prior and training data along different aspects of fine-tuning.
As in \ours{}, we use the TPE algorithm to optimize the weight hyperparameters $w_i$ to maximize the log-likelihood of the validation data $\Dval$.

Results in \cref{fig:toy} show the effect of standard hyperparameter tuning (global weight) versus considering an expanded hyperparameter space as in \ours{} (dim-wise weight).
Learning dimension-wise weights is effective, but only when the prior distribution is well-specified and provides a better signal for OOD generalization than the training set along some dimensions, as foundation models do for general fine-tuning tasks.
This additional expressivity is necessary to specify a better adaptation procedure: naively averaging the learned dim-wise parameters causes the model to overfit the ID data.
Finally, through inspecting the learned weights, we see that the learned dim-wise weights improve OOD generalization by downweighting the influence of the ID data in the directions where only the ID data differs from the prior (not the OOD data).

\section{Additional Results}
\label{sec:appendix}

\subsection{ImageNet Distribution Shifts}

\subsubsection{Main Results} \label{sec:imagenet-main}

We provide detailed results for each ImageNet-derived distribution shift in \cref{tab:imagenet-detailed}.
AutoFT outperforms all baselines on 9 out of 10 OOD datasets.
\ours{} improves OOD performance by $3.0\%$ compared to the leading baseline, FLYP.
We observe a similar effect with all other points of comparison, suggesting that perhaps these prior methods for robust fine-tuning ``overfit'' to the ID distribution in a way that \ours{} does not.



\begin{table*}[t!]
\centering
\begin{minipage}{.9\linewidth}
\small
\setlength{\tabcolsep}{3pt}
\begin{tabular}{c|c|ccccc|c|c|ccccc|c}
\toprule
& \multicolumn{7}{c}{Without Ensembling} & \multicolumn{7}{c}{With Ensembling} \\
\cmidrule(lr){1-2} \cmidrule(lr){2-8} \cmidrule(lr){9-15}
Methods & ID & Im-V2 & Im-R & Im-A & Sketch & ObjectNet & Avg. OOD & ID & Im-V2 & Im-R & Im-A & Sketch & ObjectNet & Avg. OOD \\
\midrule
Zeroshot & 68.3 & 61.9 & 77.7 & 50.0 & 48.3 & 55.4 & 58.7 & 68.3 & 61.9 & 77.7 & 50.0 & 48.3 & 55.4 & 58.7 \\
LP & 79.9 & 69.8 & 70.8 & 46.4 & 46.9 & 52.1 & 57.2 & 80.0 & 70.3 & 72.4 & 47.8 & 48.1 & 52.8 & 58.3 \\
FT & 81.3 & 71.2 & 66.1 & 37.8 & 46.1 & 53.3 & 54.9 & 82.5 & 72.8 & 74.9 & 48.1 & 51.9 & 59.0 & 61.3 \\
L2-SP & 81.7 & 71.8 & 70.0 & 42.5 & 48.5 & 56.2 & 57.8 & 82.2 & 72.9 & 75.1 & 48.6 & 51.4 & 58.9 & 61.4 \\
LP-FT & 81.7 & 72.1 & 73.5 & 47.6 & \textbf{50.3} & 58.2 & 60.3 & 82.1 & 72.8 & 75.3 & 50.1 & 51.7 & 59.2 & 61.8 \\
FLYP & \textbf{82.6} & 73.0 & 71.4 & 48.1 & 49.6 & 58.7 & 60.2 & \textbf{82.9} & 73.5 & 76.0 & 53.0 & 52.3 & 60.8 & 63.1 \\
\midrule
\ours{} & 81.8 & \textbf{73.1} & \textbf{72.4} & \textbf{48.8} & 49.8 & \textbf{63.5} & \textbf{61.5} & 82.4 & \textbf{73.6} & \textbf{76.4} & \textbf{53.1} & \textbf{52.6} & \textbf{65.6} & \textbf{64.3} \\
\bottomrule
\end{tabular}
\end{minipage}
\caption{
\textbf{Detailed ImageNet results.}
\ours{} consistently outperforms all baselines on 9 out of 10 ImageNet distribution shifts. 
}
\label{tab:imagenet-detailed}
\end{table*}

\subsubsection{Few-Shot Classification}
\begin{figure*}
\centering 
\includegraphics[width=1.0\linewidth]{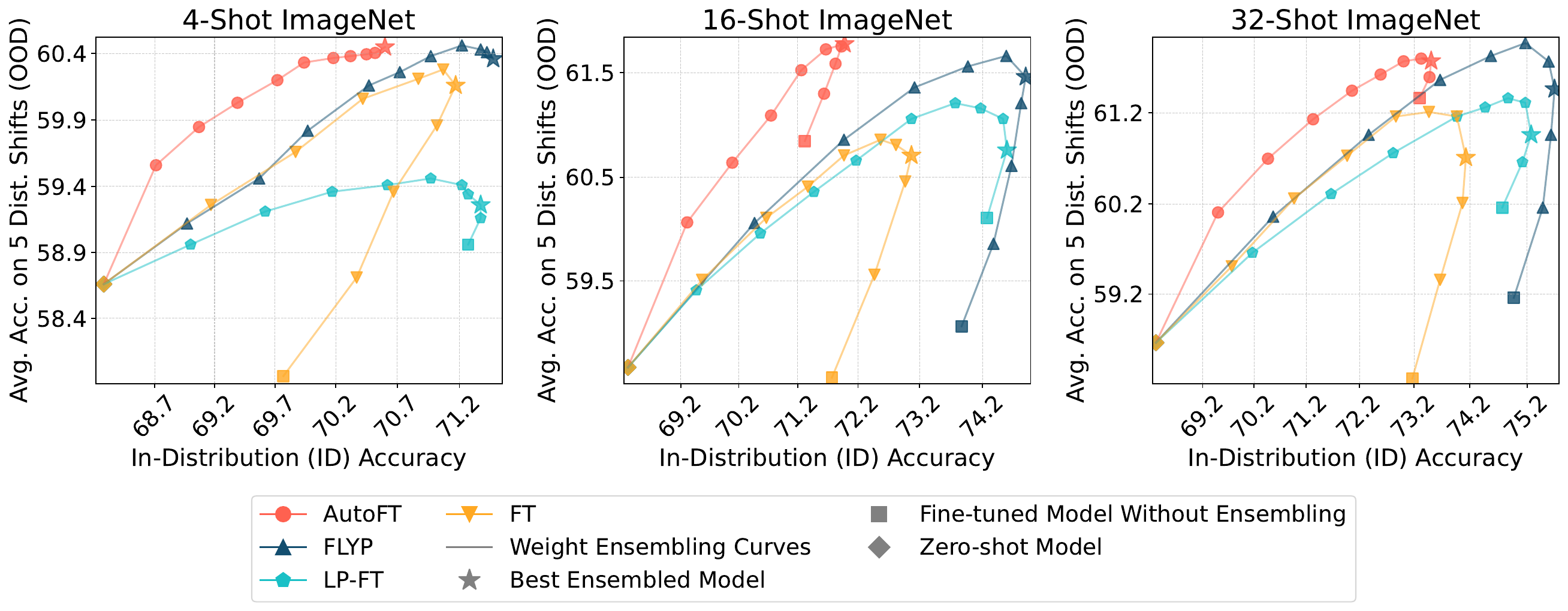}
\caption{
\textbf{\ours{} improves OOD generalization in few-shot learning.}
\ours{} results in gains in \textit{effective robustness} over baselines, indicated by vertical distance.
\ours{} results in better OOD performance at a given ID performance threshold across varying degrees of weight interpolation between the fine-tuned and zero-shot models.
}
\label{fig:imagenet-fewshot-effective-robustness-plots}
\end{figure*}

\cref{fig:imagenet-fewshot-effective-robustness-plots} plots effective robustness at varying interpolation coefficients on five ImageNet distribution shifts for 4, 16, and 32 shot classification.
In all three settings, \ours{} results in better OOD generalization and avoids overfitting to the small fine-tuning dataset.
For example, in 32-shot ImageNet, \ours{} improves average OOD accuracy by $2.2\%$ compared to the leading baseline, FLYP.

\input{tables/transfer_learning.tex}

\subsection{CIFAR-10 Distribution Shifts}
\input{tables/cifar.tex}

\ours{} also outperforms prior robust fine-tuning approaches on the subtle CIFAR-10.1 and CIFAR-10.2 distribution shifts. 
Here, \ours{} uses the CLIP ViT-L/14 backbone, following \citet{wortsman2022robust}, and $100$ examples from CIFAR-10-C to learn the fine-tuning objective and hyperparameters.

\section{Additional Experiments}
\subsection{Expressivity of Hyperparameter Space} \label{sec:expressivity}

\begin{table}
\begin{minipage}{\linewidth}
\input{tables/layerwise.tex}
\end{minipage}
\end{table}

We investigate whether more expressive hyperparameter search spaces result in better performance.
We compare \ours{} against a layerwise variant of \ours{} that learns per-layer learning rates, weight decays, and L1/L2 norms and regularization terms in \cref{tab:expressivity}.
The layerwise variant of \ours{} results in worse ID and OOD performance, suggesting that an overly expressive search space leads to overfitting of the small validation set.

\section{Experimental Details}
\label{sec:exp_details}

\subsection{Training Protocol} 
We closely follow the training details of \citet{goyal2022finetune} and \citet{wortsman2022robust}.
All methods fine-tune models with an AdamW optimizer, cosine learning rate scheduler, and a batch size of 512 for ImageNet and 256 for all other datasets. 
All baseline hyperparameters, such as learning rate, weight decay, and warmup length, are tuned through grid search.
All methods, including \ours{}, perform early stopping based on in-distribution (ID) validation accuracy.
We provide a comprehensive breakdown of the hyperparameter sweeps in the supplementary material.
We emphasize that none of these methods, including \ours{}, observe any of the test OOD distributions during training.
Finally, we report metrics averaged over 5 runs with 95\% confidence intervals.

\subsection{Baselines} We follow the training details in \citet{goyal2022finetune}.
For all datasets excluding ImageNet, we use a batch size of 256 and conduct a hyperparameter sweep across five learning rates $\{10^{-6}, 10^{-5}, \dots, 10^{-2}\}$ and five weight decay values in the range $\{0.0, 0.1, \dots, 0.4\}$.
On ImageNet, we use a larger batch size of 512 and perform a hyperparameter sweep over three learning rates $\{10^{-6}, 10^{-5}, 10^{-4}\}$ and two weight decays $\{0, 0.1\}$.
For L2-SP, we tune the weight of the regularization term $\lambda \in \{10^{-4}, 10^{-3}, 10^{-2}, 10^{-1}\}$.
We select baseline hyperparameters based on ID validation performance. 
For datasets without a standard validation set, we split the training data into an 80:20 ratio to create one. 

\subsection{\ours{}}
\label{sec:app-autoft-details}

\textbf{Hyperparameter search space.} As described in~\cref{sec:method}, \ours{} learns weights for nine different losses on a log-uniform range $[10^{-4}, 10]$.
\ours{} additionally searches for learning rate in the log-uniform range $[10^{-2} \cdot \eta^*, 10^2 \cdot \eta^*]$, where $\eta^*$ is the conventional learning rate used in prior works on fine-tuning \citep{wortsman2022robust, kumar2022fine, goyal2022finetune}, weight decay values in the log-uniform range $[0.0, 1.0]$, and random seeds in the range $[0, 100]$.
The layerwise variant of \ours{} learns per-layer learning rates and weight decay regularization terms, each sampled from the same respective range.

\textbf{WILDS-iWildCam and WILDS-FMoW state-of-the-art (SoTA).} 
For the iWildCam and FMoW SoTA results in~\cref{tab:sota_iwild}, we fine-tune the \texttt{ViT-L/14@336px} model with loss weights learned on the smaller \texttt{ViT-B/16} backbone with \ours{}.
Applying \ours{} directly to the \texttt{ViT-L/14@336px} model may improve performance further, although at the cost of more compute.

\textbf{Validation sets.} 
For the iWildCam and FMoW experiments, we use $1000$ examples from the official WILDS OOD validation splits from \citet{koh2021wilds}.
For ImageNet, we use $1000$ examples from ImageNet-C.
For CIFAR-10, we use $100$ examples from CIFAR-10-C.
For the few-shot experiments, we use a randomly sampled $k$-shot ID validation set, and average over 50 runs to account for variance in the fine-tuning and validation sets.
For the transfer learning experiments, we \textit{do not use OOD data}. 
We randomly partition the ID validation sets into two validation sets: one for hyperparameter optimization, and the other for early stopping during final fine-tuning. 
We use $100, 200, 400$ examples for Caltech-101, Stanford Cars, and Flowers-102, respectively.

\textbf{Hyper-hyperparameters.}
We list hyper-hyperparameters used in \cref{tab:hyperhyper}. 
Few-shot hyper-hyperparameters are listed in \cref{tab:hyperhyper-fewshot}.
We run \ours{} with $500$ outer-loop trials on all non-few shot experiments, which is the largest number of trials that can be run with a reasonable amount of compute.
We use grid-search to tune the number of inner steps and validation set size and select hyper-hyperparameters based on ID validation performance of the final fine-tuned model.
On iWildCam, FMoW, and CIFAR-10, we tune over $\{10, 50, 100\}$ inner steps and $\{100, 1000, 5000\}$ validation examples.
On ImageNet, we tune over $\{10, 50\}$ inner steps and $\{1000, 5000\}$ validation examples.
On Caltech-101, Stanford Cars, and Flowers-102, we tune over $\{10, 50\}$ inner steps. 
For the few-shot binary classification experiments, we run \ours{} with $50$ outer-loop evaluations and tune over $\{5, 10, 20\}$ inner steps.
For the few-shot ImageNet experiments, we run \ours{} with $100$ outer-loop evaluations and tune over $\{5, 20, 50\}$ inner steps.

\begin{table}[b]
\vspace{-3mm}
\centering
\begin{tabular}{lcc}
\toprule
\textbf{Dataset} & \textbf{Steps} & \textbf{Trials} \\
\midrule
SST-2 & 10 & 50 \\
PatchCamelyon & 10 & 50 \\
ImageNet-4 & 5 & 100 \\
ImageNet-16 & 20 & 100 \\
ImageNet-32 & 50 & 100 \\
\bottomrule
\end{tabular}
\caption{
\label{tab:hyperhyper-fewshot}
\ours{} training settings for few-shot classification experiments.
}
\end{table}

Our empirical results show that a small number of inner loop steps is effective for identifying suitable hyperparameters. 
This is also a necessity considering computational constraints. 
Our method strikes a balance between practical feasibility and reliable hyperparameter selection.

\textbf{Few-shot classification.} In the $k$-shot setting with $k \in \{4, 16, 32\}$, we select $k$ training and $k$ validation examples from each class.
To account for variance from the small training and validation sets, for each run of \ours{}, we repeat hyperparameter optimization $5$ times and select the best set of hyperparameters based on ID validation performance. 
We then average results over $50$ runs.

\textbf{Transferability of learned hyperparameters experiment details.} 
The OOD distributions for fine-tuning on CIFAR-10 are CIFAR-10.1 and CIFAR-10.2.
The OOD distributions for fine-tuning on MNIST are MNIST-C, rotated MNIST, colored MNIST, and EMNIST.

\subsection{Datasets} 
\label{sec:app-datasets}
Below, we summarize the datasets we use for evaluation, including the fine-tuning dataset (ID), the validation dataset for hyperparameter optimization, and the test OOD datasets.

\begin{itemize}
    \item \textbf{CIFAR-10} \citep{krizhevsky2009learning} contains 60,000 images across 10 classes.
    We use CIFAR-10 for fine-tuning, 100 examples from CIFAR-10-C for validation, and the CIFAR-10.1~\citep{recht2018cifar10.1,torralba2008tinyimages} and CIFAR-10.2~\citep{lu2020harder} as OOD test sets.

    \item \textbf{ImageNet}~\citep{deng2009imagenet} contains over a million images in 1000 categories. 
    We use ImageNet as our ID distribution, 15000 examples from ImageNet-C for validation, and five ImageNet variations for the OOD datasets following prior works~\citep{clip,wortsman2022robust,kumar2022fine,goyal2022finetune}: ImageNet-V2~\citep{recht2019imagenet}, ImageNet-R~\citep{hendrycks2021many}, ImageNet-A~\citep{hendrycks2021natural}, ImageNet-Sketch~\citep{wang2019learning}, and ObjectNet \citep{barbu2019objectnet}.
    
    \item \textbf{WILDS-iWildCam} \citep{beery2021iwildcam, koh2021wilds, sagawa2022extending} is an animal camera trap image classification dataset where the label $y$ is one of 182 animal species.
    The ID and OOD datasets consist of photos from disjoint sets of camera traps, making the images differ in camera specifications and attributes such as background and lighting.
    We use the official splits from~\citet{koh2021wilds} and use the ID train set for fine-tuning, the OOD validation set for hyperparameter optimization, and the OOD test set for evaluation.

    \item \textbf{WILDS-FMoW} \citep{christie2018functional, koh2021wilds, sagawa2022extending} contains remote sensing imagery from satellites. 
    Each image is to be classified into one among 62 categories, including labels like ``impoverished settlement'' and ``hospital.''
    The ID and OOD datasets differ in year of acquisition and geographic location.
    We use the official splits from~\citet{koh2021wilds} and use the ID train set for fine-tuning, the OOD validation set for hyperparameter optimization, and the OOD test set for evaluation.
\end{itemize}

In all of the transfer learning datasets described below, we use a subset of the ID validation set for hyperparameter optimization. 
In other words, we do not use an ``OOD'' validation set.
\begin{itemize}
    \item \textbf{Caltech101} \citep{fei2004learning} contains images of objects from 101 different categories, including ``dragonfly,'' ``grand piano,'' and ``saxophone.''

    \item \textbf{StanfordCars} \citep{krause20133d} features a collection of car images categorized by model, make, and year, where the task is to classify them into one of 196 types, such as ``Ford Mustang Convertible 1967'' or ``Toyota Prius Hatchback 2009.''

    \item \textbf{Flowers102} \citep{veeling2018rotation} consists of flower images from the UK, with the objective of classifying each image into one of 102 species, such as ``oxeye daisy'' or ``hibiscus.''

    \item \textbf{PatchCamelyon} \citep{veeling2018rotation} provides digital pathology images for binary classification, with the goal of identifying metastatic tumor tissues.

    \item \textbf{Rendered SST2} \citep{radford2021learning} is a dataset for optical character recognition, where the task is to classify text sentiment as ``positive'' or ``negative.''
\end{itemize}

\label{sec:appendix-hyperparams}

%% file: tables/transfer_learning.tex

\begin{table*}[t!]
\vspace{-2mm}
\centering
\begin{tabular}{lccc}
\toprule
Methods & CalTech-101 & Stanford Cars & Flowers-102 \\
\midrule
Zeroshot & 87.7 (-) & 64.4 (-) & 71.2 (-) \\
LP & 94.8 (0.0) & 83.1 (0.0) & 95.9 (0.0) \\
FT & 97.2 (0.1) & 84.4 (0.3) & 90.4 (0.5) \\
LP-FT & 96.9 (0.6) & 89.4 (0.1) & \textbf{97.9 (0.1)} \\
FLYP & 97.6 (0.1) & \textbf{89.6 (0.3)} & 97.7 (0.1) \\
\midrule
\oursbf{} & \textbf{99.0 (0.1)} & \textbf{89.6 (0.2)} & 97.6 (0.2) \\
\bottomrule
\end{tabular}
\vspace{-2mm}
\caption{
\textbf{\ours{} can lead to improvements in IID transfer learning.} In addition to enhancing OOD generalization, \ours{} can improve performance in IID transfer learning settings when using an ID validation set to learn the objective and hyperparameters..
}
\label{tab:transfer_learning}
\end{table*}

%% file: tables/cifar.tex
\begin{table}
\centering
\begin{tabular}{ccc}
\toprule
Method & CIFAR-10.1 & CIFAR-10.2 \\
\midrule
Zero-shot & 92.5 & 88.8 \\
Fine-tuning & 95.9 & 91.3 \\
\oursbf & \textbf{97.5} & \textbf{93.5} \\
\midrule
WiSE-FT (best $\alpha$) & 98.0 & 94.4 \\
\oursbf{} (best $\alpha$) & \textbf{98.3} & \textbf{95.0} \\
\bottomrule
\end{tabular}
\caption{\textbf{Evaluation on CIFAR-10 distribution shifts.} \ours{} outperforms fine-tuning by $2.2\%$ on CIFAR-10.2 and by $1.4\%$ on CIFAR-10.1, using only 100 samples from CIFAR-10-C.
\ours{} additionally outperforms WiSE-FT with weight ensembling.}
\label{tab:cifar10}
\vspace{-5.0pt}
\end{table}

%% file: tables/layerwise.tex

\centering
\begin{tabular}{ccc}
\toprule
\multicolumn{3}{c}{iWildCam} \\
\midrule 
Method & ID & OOD \\
\midrule
\ours & \textbf{51.0} & \textbf{38.3} \\
Layerwise \ours & 47.8 & 34.6 \\
\bottomrule
\end{tabular}
\caption{\textbf{Expressivity of the \ours{} search space.} \ours{} results in better performance than its layerwise variant with per-layer learning rates, weight decays, and L1/L2 distances to pretrained parameters. Evaluation is on the iWildCam benchmark. This suggests that more expressive hyperparameter spaces overfit to the validation set, degrading performance.}
\label{tab:expressivity}
\vspace{-5.0pt}